\documentclass[10pt,twocolumn]{article}

\usepackage{cvpr}
\usepackage{times}
\usepackage{epsfig}
\usepackage{graphicx}
\usepackage{amsmath}
\usepackage{amssymb}

\usepackage{enumitem}
\usepackage{multirow}
\usepackage{algorithm}
\usepackage{algorithmicx}
\usepackage{algpseudocode}
\usepackage{appendix}

\usepackage{footmisc}
\usepackage[dvipsnames]{xcolor}
\usepackage{colortbl,booktabs}
\usepackage{threeparttable}
\usepackage{nopageno}
\usepackage{comment}
\usepackage{url}
\usepackage{algpseudocode}
\usepackage[small,bf]{caption}
\usepackage{mathtools}
\usepackage{pifont}
\usepackage{lipsum}
\usepackage{adjustbox}
\usepackage{booktabs}       
\usepackage{amsfonts}       
\usepackage{nicefrac}       
\usepackage{microtype}      
\usepackage{makecell}
\usepackage{diagbox}

\newcommand{\red}[1]{\textcolor{red}{#1}}

\usepackage[pagebackref=true,breaklinks=true,colorlinks,bookmarks=false]{hyperref}

\cvprfinalcopy 

\def\cvprPaperID{5582} 

\ifcvprfinal\fi
\begin{document}

\title{L$^2$-GCN: Layer-Wise and Learned Efficient Training of \\ Graph Convolutional Networks}

\author{Yuning You\footnotemark[1],\, Tianlong Chen\footnotemark[1],\, Zhangyang Wang, \,Yang Shen\\
Texas A\&M University\\
{\tt\small \{yuning.you,wiwjp619,atlaswang,yshen\}@tamu.edu}
}

\maketitle

\begin{abstract}
Graph convolution networks (GCN) are increasingly popular in many applications, yet remain notoriously hard to train over large graph datasets. They need to compute node representations recursively from their neighbors. Current GCN training algorithms suffer from either high computational costs that grow exponentially with the number of layers, or high memory usage for loading the entire graph and node embeddings. In this paper, we propose a novel efficient layer-wise training framework for GCN (L-GCN), that disentangles feature aggregation and feature transformation during training, hence greatly reducing time and memory complexities. We present theoretical analysis for L-GCN under the graph isomorphism framework, that L-GCN leads to as powerful GCNs as the more costly conventional training algorithm does, under mild conditions. We further propose L$^2$-GCN, which learns a controller for each layer that can automatically adjust the training epochs per layer in L-GCN. Experiments show that L-GCN is faster than state-of-the-arts by at least an order of magnitude, with a consistent of memory usage not dependent on dataset size, while maintaining comparable prediction performance.
With the learned controller, L$^2$-GCN can further cut the training time in half.
Our codes are available at \url{https://github.com/Shen-Lab/L2-GCN}.
\end{abstract}
\renewcommand{\thefootnote}{\fnsymbol{footnote}}
\footnotetext[1]{Equal Contribution.}
\begin{figure}[ht]
\begin{center}
  \includegraphics[width=1\linewidth]{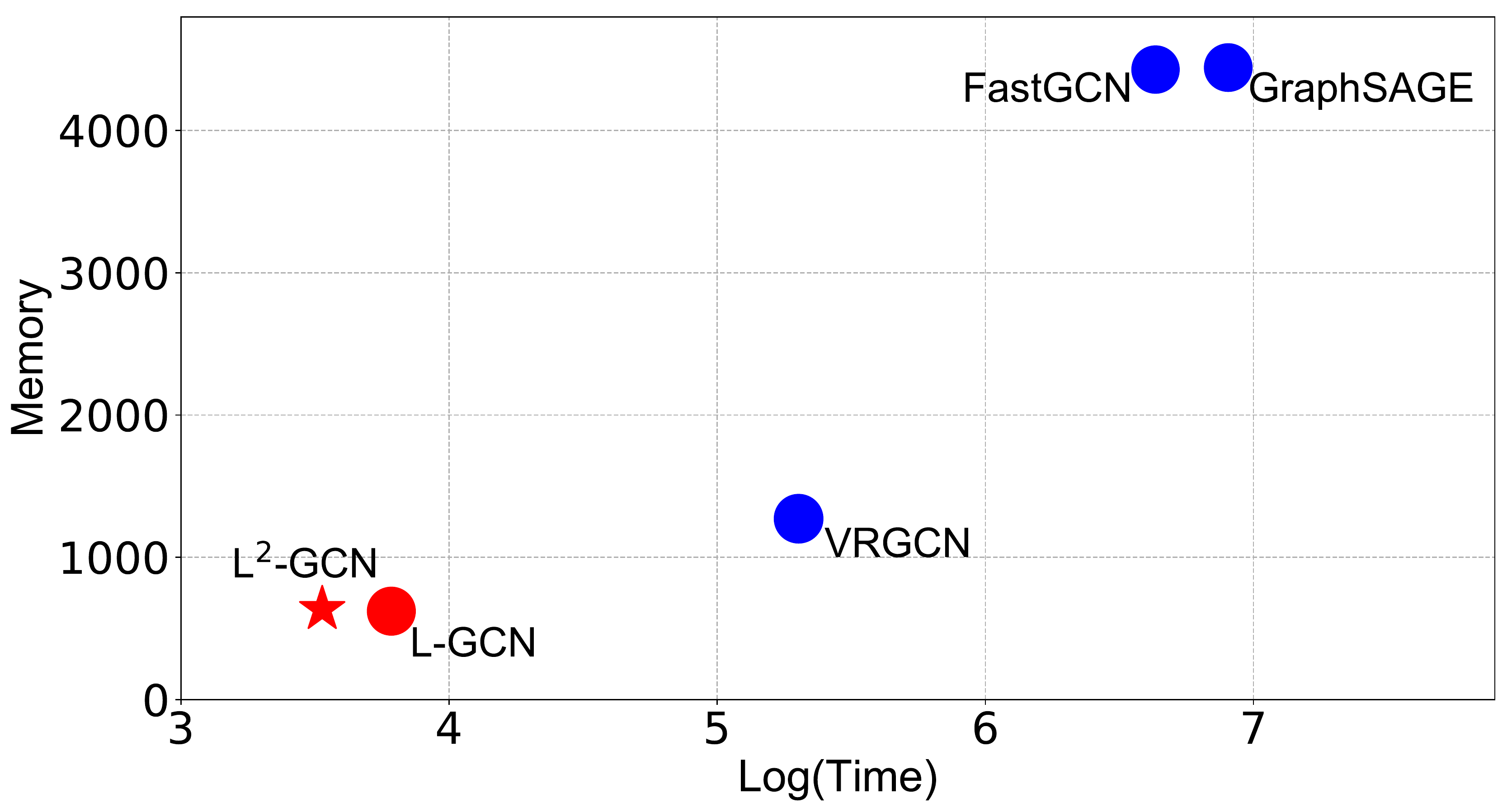}
\end{center}
  \caption{Summary of our achieved performance and efficiency on Reddit. \textbf{The lower left corner indicates the desired lowest complexity in time (training time) and memory consumption (GPU memory usage).} The size of markers represents F1 scores. Blue circles (\textcolor{blue}{$\bullet$}) are state-of-the-art mini-batch training algorithms, red circle (\textcolor{red}{$\bullet$}) is L-GCN, and red star (\textcolor{red}{\ding{72}}) is L$^2$-GCN.
  The corresponding F1 scores are: GraphSAGE (\textcolor{blue}{$\bullet$}, 93.4), FastGCN (\textcolor{blue}{$\bullet$}, 92.6), VRGCN (\textcolor{blue}{$\bullet$}, 96.0), L-GCN (\textcolor{red}{$\bullet$}, 94.2) and L$^2$-GCN(\textcolor{red}{\ding{72}}, 94.0).}
\label{fig:intro}
\end{figure}


\begin{table*}[t]
\small
\begin{center}
\caption{Time and memory complexities: we consider time complexity for the feature propagation in the network, and memory complexity for storing node embeddings in each layer (since given the same network architecture, the weight storage costs the same memory for all compared algorithms c). $L$ is the layer number, $D$ the feature dimension, $N$ the node number, $S_\mathrm{NEI}$ the neighborhood size, $B$ the minibatch size, $S$ the training sample size, $S_\mathrm{VR}$ the reduced sample size, and $N_\mathrm{BAT}$ the minibatch number.}
\label{tab:complexity}
\resizebox{1\textwidth}{!}{
\begin{tabular}{c c c c c c c}
\hline
 & GCN \cite{kipf2016semi} & Vanilla Mini-Batch & GraphSAGE \cite{hamilton2017inductive} & FastGCN \cite{chen2018fastgcn}& VRGCN \cite{chen2017stochastic} & \textbf{L-GCN} \\
\hline
Time & $O(L ||\hat{\boldsymbol{A}}||_0 D + L N D^2)$ & $O(S^L_\mathrm{NEI} B D^2)$ & $O(S^L B D^2)$ & $> O(S L B D^2)$ & $> O(S^L_\mathrm{VR} B D^2)$ & $O(L \frac{||\hat{\boldsymbol{A}}||_0}{N_\mathrm{BAT}} D + B D^2)$ \\
Memory & $O(L N D)$ & $O(S^L_\mathrm{NEI} B D)$ & $O(S^L B D)$ & $> O(S L B D)$ & $O(L N D)$ & $O(B D)$ \\
 \hline
\end{tabular}}
\end{center}
\end{table*}


\section{Introduction}
Graph convolution networks (GCN) \cite{kipf2016semi} generalize convolutional neural networks (CNN) \cite{lecun1995convolutional} to graph data. Given a node in a graph, a GCN first \textit{aggregates} the node embedding with its neighbor node embeddings, and then \textit{transforms} the embedding through (hierarchical) feed-forward propagation. The two core operations, \ie, aggregating and transforming node embeddings, take advantage of the graph structure and outperform structure-unaware alternatives \cite{perozzi2014deepwalk, tang2015line, grover2016node2vec}. GCNs hence demonstrate prevailing success in many graph-based applications, including node classification \cite{kipf2016semi}, link prediction \cite{zhang2018link} and graph classification \cite{ying2018graph}. 


However, the training of GCNs has been a headache, and a hurdle to scale up GCNs further. How to train CNNs more efficiently has recently become a popular topic of explosive interest, by bypassing unnecessary data or reducing expensive operations \cite{wang2019e2,you2019drawing,jiang2019accelerating}. For GCNs, as the graph dataset grows, the large number of nodes and the potentially dense adjacency matrix prohibit fitting them all into the memory, thus putting \textit{full-batch} training algorithms (\ie, those requiring the full data and holistic adjacency matrix to perform) in jeopardy. That motivates the development of \textit{mini-batch} training algorithms, \ie, treating each node as a data point and updating locally. In each mini-batch, the embedding of a node at the $l$th layer is computed from the neighborhood node embeddings at the ($l-1$)-th layer through the graph convolution operation. As the computation is performed recursively through all layers, 
the mini-batch complexity will increase exponentially with respect to the layer number. To mitigate the complexity explosion, several sampling-based strategies have been adopted, \eg GraphSAGE \cite{hamilton2017inductive} and FastGCN \cite{chen2018fastgcn}, yet with few performance guarantees. VRGCN \cite{chen2017stochastic} reduces the sample size through variance reduction, and guarantees its performance convergence to the full-sample approach, but it requires to store the full-batch node embeddings of each layer in the memory, limiting its efficiency gain. Cluster-GCN \cite{chiang2019cluster} used graph clustering to partition the large graph into subgraphs, and performs subgraph-level mini-batch training, yet again being only empirical.

In this paper, we propose a novel layer-wise training algorithm for GCNs, called  (\textbf{L-GCN}). The key idea is to decouple the two key operations in the per-layer feedforward graph convolution: feature aggregation (FA) and feature transformation (FT), whose concatenation and cascade result in the exponentially growing complexity. Surprisingly, the resulting greedy algorithm will not necessarily compromise the network representation capability, as shown by our theoretical analysis inspired by \cite{xu2018powerful} using a graph isomorphism framework. To bypass extra hyper-parameter tuning, we then introduce layer-wise and learned GCN training (\textbf{L$^2$-GCN}), which learns a controller for each layer that can automatically adjust the training epochs per layer in L-GCN. Table \ref{tab:complexity} compares the training complexity between L-GCN, L$^2$-GCN and existing competitive algorithms, demonstrating our approaches' remarkable advantage in reducing both time and memory complexities. More experiments show that our proposed algorithms are significantly faster than state-of-the-arts, with a consistent usage of GPU memory not dependent on dataset size, while maintaining comparable prediction performance. 
Our contributions can be summarized below:
\begin{itemize}
    \item A layer-wise training algorithm for GCNs with much lower time and memory complexities;
    \item Theoretical justification that under some sufficient conditions
    the greedy algorithm does not compromise in the graph-representative power;
    \item Learned controllers that automatically configure layer-wise training epoch numbers, in place of manual hyperparameter tuning;
    \item State-of-the-art performance achieved in addition to the light weight, on extensive applications.
\end{itemize}

\section{Related Work}
We follow \cite{chiang2019cluster} to categorize existing GCN training algorithms into \textit{full-batch} and \textit{mini-batch} (stochastic) algorithms, and compare their pros and cons.


\begin{figure*}[t]
\begin{center}
  \includegraphics[width=0.8\linewidth]{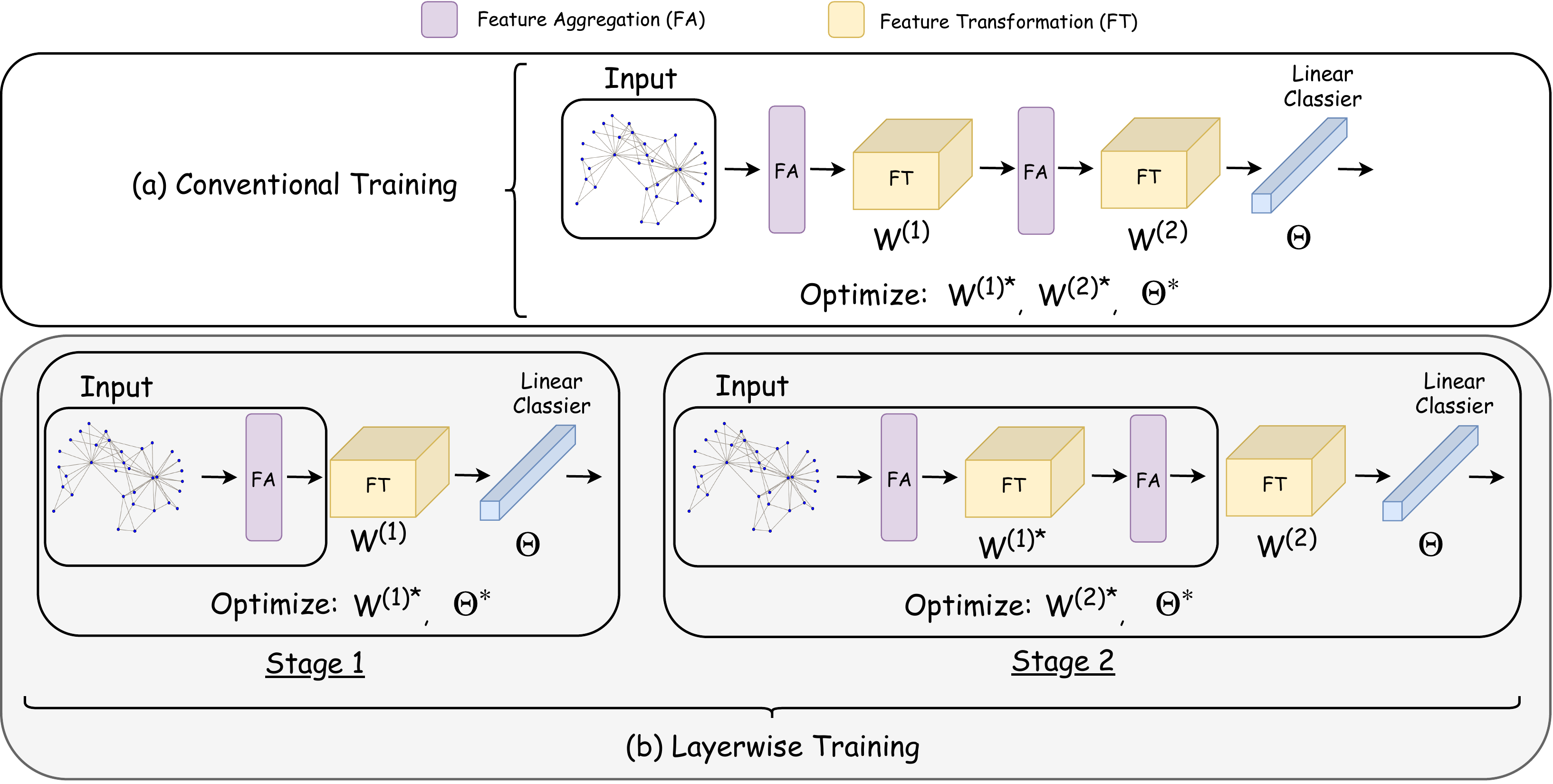}
\end{center}
  \caption{Conventional training vs layer-wise training for a two-layer GCN. (a) In conventional training, the optimizer jointly optimizes the weight matrix $\boldsymbol{W}^{(1)}, \boldsymbol{W}^{(2)}$ as $\boldsymbol{W}^{(1)*}, \boldsymbol{W}^{(2)*}$ and the linear classifier $\boldsymbol{\Theta}$ as $\boldsymbol{\Theta}^*$, saving $\boldsymbol{W}^{(1)*}, \boldsymbol{W}^{(2)*}, \boldsymbol{\Theta}^*$ for the network. (b) In layer-wise training, two layers are trained in two sequential stages: in the first stage the optimizer optimizes the weight matrix $\boldsymbol{W}^{(1)}$ as $\boldsymbol{W}^{(1)*}$ and the linear classifier $\boldsymbol{\Theta}^{(1)}$ as $\boldsymbol{\Theta}^{(1)*}$, saving $\boldsymbol{W}^{(1)*}$ for the second stage and dropping $\boldsymbol{\Theta}^{(1)*}$; then in the second stage the optimizer optimizes the weight matrix $\boldsymbol{W}^{(2)}$ as $\boldsymbol{W}^{(2)*}$ and the linear classifier $\boldsymbol{\Theta}^{(2)}$ as $\boldsymbol{\Theta}^*$, saving $\boldsymbol{W}^{(2)*}, \boldsymbol{\Theta}^*$ for the network.}
\label{figure1}
\end{figure*}

\subsection{Full-Batch GCN Training}
The original GCN \cite{kipf2016semi} adopted the full-batch gradient descent algorithm. 
Let's define an undirected graph as 
$G = (V, E)$
, where $V = \{v_1,...,v_N\}$ 
represents the vertex set with $N$ nodes, and $E = \{e_1,...,e_{N_E}\}$ represents the edge set with $N_E$ edges: $e_n = (v_i, v_j)$ indicates an edge between vertices $v_i$ and $v_j$. $\boldsymbol{F} \in \mathbb{R}^{N \times D}$ is the feature matrix with the feature dimension $D$, and $\boldsymbol{A} \in \mathbb{R}^{N \times N}$ is the adjacency matrix where $a_{ij} = a_{ji} = \{ \begin{smallmatrix} 1, \, \text{if} \, (v_i, v_j) \, \text{or} \, (v_j, v_i) \in E \\ 0, \, \text{otherwise} \end{smallmatrix} $. By constructing an $L$-layer GCN, we express the output $\boldsymbol{X}^{(l)} \in \mathbb{R}^{N \times D^{(l)}}$ of the $l$th layer and the network loss as:
\begin{equation} \label{eq1}
\boldsymbol{X}^{(l)} = \sigma( \hat{\boldsymbol{A}} \boldsymbol{X}^{(l-1)} \boldsymbol{W}^{(l)}), \text{Loss} = Loss(\boldsymbol{X}^{(L)}, \boldsymbol{\Theta}, \boldsymbol{Y}),
\end{equation}
where $\hat{\boldsymbol{A}}$ is the regularized adjacency matrix, $\boldsymbol{X}^{(0)} = \boldsymbol{F}$, $\boldsymbol{W}^{(l)} \in \mathbb{R}^{D^{(l-1)} \times D^{(l)}}$ is the weight matrix, $D^{(0)} = D$, $\sigma(\cdot)$ is a nonlinear function, $\boldsymbol{\Theta} \in \mathbb{R}^{D^{(L)} \times D_{CLA}}$ is the linear classification matrix, $\boldsymbol{Y}$ the training labels, and $Loss(\cdot)$ the loss function. For simplicity and without affecting the analysis, we set $D^{(1)} = ... = D^{(L)} = D$.

For the time complexity of the network propagation in \eqref{eq1}, $\hat{\boldsymbol{X}}^{(l)} = \hat{\boldsymbol{A}} \boldsymbol{X}^{(l-1)}$ costs $O(||\hat{\boldsymbol{A}}||_0 D)$ in time and $\hat{\boldsymbol{X}}^{(l)} \boldsymbol{W}^{(l)}$ costs $O(N D^2)$ in time, which in total leads to $O(L ||\hat{\boldsymbol{A}}||_0 D + L N D^2)$ time consumed for the entire network. For the memory complexity, storing the $L$-layer embeddings $\boldsymbol{X}^{(l)}, l = 1, ..., L$ requires $O(L N D)$ in memory. Both time and memory complexities are proportional to $N$, which cannot scale up well for large graphs.

\subsection{Mini-Batch SGD Algorithms}
The vanilla mini-batch SGD algorithm propagates the vertex representations in a minibatch, rather than for all nodes. We rewrite the network propagation \eqref{eq1} for the $i$th node in the $l$th layer as:
\begin{equation} \label{eq4}
\begin{split}
& \boldsymbol{x}^{(l)}_i = \sigma( ( \hat{a}_{ii} \boldsymbol{x}^{(l-1)}_i + \sum_{j = 1, ..., N,\ \text{s.t.}\ \hat{a}_{ij} \neq 0} a_{ij} \boldsymbol{x}^{(l-1)}_j ), \boldsymbol{W}^{(l)} ), \\ &\medspace \text{Loss} = \frac{1}{N} \sum^N_{i=1} Loss(\boldsymbol{x}^{(L)}_i, \boldsymbol{\Theta}, \boldsymbol{Y}),
\end{split}
\end{equation}
where $\boldsymbol{x}^{(0)}_i = F[i, :]$. With \eqref{eq4}, we can feed the feature matrix $\boldsymbol{F}$ in a mini-batch dataloader and run the stocastic gradient descent (SGD) optimizer. Suppose $B$ is the minibatch size and $S_\mathrm{NEI}$ the neighborhood size, the time complexity for the propagation per mini-batch is $O(S^L_\mathrm{NEI} B D^2)$ and the memory complexity is $O(S^L_\mathrm{NEI} B D)$. We next discuss a few variants on top of the vanilla minim-batch algorithm:
\begin{itemize}
\item \textbf{GraphSAGE \& FastGCN.} \cite{hamilton2017inductive, chen2018fastgcn} \medspace
Both adopted sampling scheme to reduce complexities. GraphSAGE proposes to use fixed-size sampling for the neighborhood in each layer. It yet suffers from the ``neighborhood expansion" problem, making its time and memory complexities grow exponentially with the layer number. FastGCN proposes global importance sampling rather than local neighborhood sampling, alleviating the complexity growth issue. Suppose $S \le S_\mathrm{NEI}$ is the sample size, the time and memory complexities are $O(S^L B D^2)$ and $O(S^L B D)$ for GraphSAGE, and $O(S L B D^2)$ and $O(S L B D)$ for FastGCN, respectively. What's more, \cite{zou2019layer} develops layer-dependent importance sampling based on FastGCN and further achieves both time and memory efficiency. Besides, for FastGCN, there is extra complexity requirement for importance weight computation.
\item \textbf{VRGCN.} \cite{chen2017stochastic} \medspace
proposes to use variance reduction to reduce the sample size in each layer, which managed to achieve good performance with smaller graphs. Unfortunately, it requires to store all the vertex intermediate embeddings during training, which leads to its memory complexity coming close to the full-batch training. Suppose $S_\mathrm{VR} \le S$ is the reduced sample size, the time and memory complexities of VRGCN are $O(S^L_\mathrm{VR} B D^2)$ and $O(L N D)$, respectively (plus some overhead for computing variance reduction).
\item \textbf{Cluster-GCN.} \cite{chiang2019cluster} \medspace
Instead of feeding nodes and their neighbors directly, \cite{chiang2019cluster} first uses a graph clustering algorithm to partition subgraphs, and then runs the SGD optimizer over each subgraph. The performance of this approach heavily hinges on the chosen graph clustering algorithm. It is further difficult to ensure training stability, \eg, w.r.t different clustering settings. 
\end{itemize}


\section{Proposed Algorithm}
To discuss the bottleneck of graph convolutional network (GCN) training algorithms, we first analyze the propagation of GCN following \cite{wu2019simplifying} and factorize the propagation \eqref{eq1} into \textit{feature aggregation (FA)} and \textit{feature transformation (FT)}.

\textbf{Feature aggregation.} To learn the node representation $\boldsymbol{X}^{(l)}$ of the $l$th layer, in the first step GCN follows the neighborhood aggregation strategy, where in the $l$th layer it updates the representation of each node by aggregating the representations of its neighbors, and at the same time the representation of itself is aggregated by the representations of its neighbors, which is written as:
\begin{equation} \label{eq2}
\hat{\boldsymbol{X}}^{(l)} = \hat{\boldsymbol{A}} \boldsymbol{X}^{(l-1)}.
\end{equation}
With \eqref{eq2}, the time and memory complexity is highly dependent on the edge number, and in the mini-batch SGD algorithm it is highly dependent on the sample size. Since during mini-batch SGD training for an $L$-layer network, $L$ times of FA for each node requires its $L$-th order neighbor nodes' representations, which results in sampling a large number of neighbor nodes. FA is the main barrier for reducing the time and memory complexity of GCN in the mini-batch SGD algorithm.

\textbf{Feature transformation.} After FA, in the second step GCN conducts FT in the $l$th layer, which consists of linear and nonlinear transformations:
\begin{equation} \label{eq3}
\boldsymbol{X}^{(l)} = \sigma( \hat{\boldsymbol{X}}^{(l)} \boldsymbol{W}^{(l)}).
\end{equation}
With \eqref{eq3}, the complexity is mainly relevant to the feature dimension. $L$ times FT for a node only requires its own representation in each layer. Given the supervised node labels $\boldsymbol{Y}$, the conventional training process for a GCN is formulated as:
\begin{equation} \label{conventional_training}
\begin{split}
& (\boldsymbol{W}^{(1)*}, ..., \boldsymbol{W}^{(L)*}, \boldsymbol{\Theta}^*) = \\
& \min_{\boldsymbol{W}^{(1)}, ..., \boldsymbol{W}^{(L)}, \boldsymbol{\Theta}} \medspace Loss(\boldsymbol{X}^{(L)}, \boldsymbol{\Theta}, \boldsymbol{Y}), \medspace s.t. \medspace \eqref{eq2},\eqref{eq3}.
\end{split}
\end{equation}

For the entire propagation of a mini-batch SGD over an $L$-layer GCN, there are $L$ times of FA and FT in each batch as shown in Figure~\ref{figure1}(a) . Without FT, $L$ times of FA can aggregate the structure information, which lacks representation learning and is still time- and memory-consuming. Without FA, $L$ times of FT is no more than a multi-layer perceptron (MLP), which efficiently learns the representation but lacks structure information. 

\subsection{L-GCN: Layer-wise GCN Training}
As described earlier, the one-batch propagation of the conventional training for an $L$-layer GCN consists of $L$ times of feature aggregation (FA) and feature transformation (FT).  Both FA and FT are necessary for capturing graph structures and learning data representations but the coupling between the two leads to inefficient training.  We therefore propose a \textit{layer-wise training algorithm} (L-GCN) to properly separate the FA and FT processes while training GCN layer by layer.

We illustrate the L-GCN algorithm in Figure~\ref{figure1}(b). For training the $l$th layer, we do FA once for all the $(l-1)$th vertex representations, aggregating its $l$th order structure information, and then feed the vertex embeddings into a single layer perceptron and run the mini-batch SGD optimizer for batches. The $l$th layer is trained by solving
\begin{equation} \label{layered_training}
\begin{split}
& (\boldsymbol{W}^{(l)*}, \boldsymbol{\Theta}^*) = \\
& \min_{\boldsymbol{W}^{(l)}, \boldsymbol{\Theta}} \medspace Loss\left(\sigma( \hat{\boldsymbol{A}} \boldsymbol{X}^{(l-1)} \boldsymbol{W}^{(l)}), \boldsymbol{\Theta}, \boldsymbol{Y}\right).
\end{split}
\end{equation}

Note that $\boldsymbol{X}^{(l-1)}$ depends on $(\boldsymbol{W}^{(1)*},\ldots,\boldsymbol{W}^{(l-1)*})$.  After finishing the $l$th layer training, we save the weight matrix  $\boldsymbol{W}^{(l)*}$ between the current input layer and hidden layer as the weight matrix of the $l$th layer, drop the weight matrices between hidden layer and output layer unless $l = L$, and calculate the $l$th-layer representations. This process is repeated until all layers are trained. 

The time and memory complexities are significantly lower compared to the conventional training and the state of the arts, as shown in Table \ref{tab:complexity}. For the time complexity, L-GCN only conducts FA $L$ times in the entire training process and FT does once per batch, whereas the conventional mini-batch training conducts FA $L$ times and FT $L$ times in each batch.  Suppose that the total training batch number is $N_\mathrm{BAT}$, the time complexity of L-GCN is $O(L \frac{||\hat{\boldsymbol{A}}||_0}{N_{BAT}} D + B D^2)$. The memory complexity is $O(B D)$ since L-GCN only trains a single layer perceptron in each batch.

\subsection{Theoretical Justification of L-GCN}
We set out to answer the following question theoretically for L-GCN: \textit{How close could the performance of layer-wise trained GCN be compared with conventionally trained GCN?} To establish the theoretical background of our layer-wise training algorithm, we follow Xu and coworker's work \cite{xu2018powerful} and show that a layer-wise trained GCN can be as powerful as a conventionally trained GCN under certain conditions. 

In \cite{xu2018powerful}, the representation power of an aggregation-based graph neural network (GNN) is evaluated, when input feature space is countable, as the ability to map any two different nodes into different embeddings. The evaluation of the representation power is extended to the ability to map any two non-isomorphic graphs into non-isomophic embeddings, where the graphs are generated as the rooted subtrees of the corresponding nodes. 
An $L$-layer GNN $\mathcal{A}: \mathbb{G} \rightarrow \mathbb{R}^D$ (excluding the linear classifier described earlier) can be represented \cite{xu2018powerful} as:
\begin{equation} \label{L_layer_GNN}
\mathcal{A} = \mathcal{R} \circ \mathcal{L}^{(L)} \circ ... \circ \mathcal{L}^{(1)},
\end{equation}
where $\mathcal{L}^{(l)}: \mathbb{R}^D \times \mathbb{M}^D \rightarrow \mathbb{R}^D, (l=1,\ldots,L)$ is the vertex-wise aggregating mapping
, $\mathbb{M}^D$ is the multiset of dimension $D$, and $\mathcal{R}: \mathbb{M}^D \rightarrow \mathbb{R}^D$ is the readout mapping as: \begin{equation} \label{layer_and_readout_mapping}
\begin{split}
& \boldsymbol{x}^{(l)}_i = \mathcal{L}^{(l)}( \boldsymbol{x}^{(l-1)}_i, \{ \boldsymbol{x}^{(l-1)}_j: j \in \mathcal{N}(i) \} ), l \in L, \\
& \text{Output} = \mathcal{R}( \{ \boldsymbol{x}^{(L)}_i: i \in N \} ),
\end{split}
\end{equation}
where $\mathcal{N}(i)$ is the set of node neighbors for the $i^{th}$ node.

Since GCN belongs to aggregation-based GNN, we use the same graph isomorphism framework for our analysis. Xu et al. \cite{xu2018powerful} provided the upper-bound power of GNN as Weisfeiler-Lehman graph isomorphism test (WL test) \cite{weisfeiler1968reduction}, and proved sufficient conditions for GNN to be as powerful as the WL test, which is described in the following lemma and theorem.

\textbf{Lemma 1.} \cite{xu2018powerful} Let $G_1$ and $G_2$ be any two non-isomorphic graphs, i.e. $G_1 \ncong G_2$. If a GNN $\mathcal{A}$ maps $G_1$ and $G_2$ into different embeddings, the WL test also decide $G_1$ and $G_2$ are not isomorphic.

\textbf{Theorem 2.} \cite{xu2018powerful} Let $\mathcal{A} = \mathcal{R} \circ \mathcal{L}^{(L)} \circ ... \circ \mathcal{L}^{(1)}$ be a GNN with sufficient number of GNN layers, $\mathcal{A}$ maps any graphs $G_1$ and $G_2$ that the WL test of isomorphism decides as non-isomorphic, to different embeddings if the following conditions hold: a) The mappings $\mathcal{L}^{(l)}, l \in L$ are injective. b) The readout mapping $\mathcal{R}$ is injective.

We further propose to use the graph isomorphism framework to characterize the ``power" of a GNN. In this framework, we observe the fact that for an aggregation-based GNN $\mathcal{A}$ (such as GCN), with any pair of isomorphic graphs $G_1$ and $G_2$, we always have $\mathcal{A}(G_1) = \mathcal{A}(G_2)$ due to the identical input and aggregation-based mapping.  In contrast, for any pair of non-isomorphic graphs $G_1$ and $G_2$, there exists certain probability $\epsilon$ that $\mathcal{A}$ wrongly maps them into identical embeddings, i.e. $\mathcal{A}(G_1) = \mathcal{A}(G_2)$, as shown in Table \ref{tab:cond_prob}. Therefore, to further analyze our algorithm, we first define a specific metric to evaluate the capacity of a GNN, as the probability of mapping any non-isomorphic graphs into different embeddings.

\begin{table}[ht]
\small
\begin{center}
\caption{Conditional probabilities of GNN $\mathcal{A}$ identifying two graphs $G_1$ and $G_2$ given their (non)isomorphism.}
\label{tab:cond_prob}
\begin{tabular}{c c c}
\hline
 & $G_1 \cong G_2$ & $G_1 \ncong G_2$ \\
\hline
$\mathcal{A}(G_1) = \mathcal{A}(G_2)$ & 1 & $\epsilon$ \\
$\mathcal{A}(G_1) \neq \mathcal{A}(G_2)$ & 0 & $\boldsymbol{1 - \epsilon}$ \\
\hline
\end{tabular}
\end{center}
\end{table}

\textbf{Definition 3.} Let $\mathcal{A}$ be a GNN; $G_1$ and $G_2$ are i.i.d. The capacity of $\mathcal{A}$, $C_{\mathcal{A}}$, is defined as the probability to map $G_1$ and $G_2$ into different embeddings if they are non-isomorphic:
\begin{equation} \label{eq5}
C_{\mathcal{A}} \equiv \mathrm{Prob}\{ \mathcal{A}(G_1) \neq \mathcal{A}(G_2) | G_1 \ncong G_2 \}.
\end{equation}

\noindent Higher capacity of a GNN indicates its  stronger distinguishing capability between non-isomorphic graphs, which corresponds to more power in graph isomorphism framework. In other words, not so powerful network will have a higher probability to map non-isomorphic graphs into the same embeddings and fail to distinguish them. With Theorem 2 and Definition 3, we have $C_{\mathcal{A}} \le C_{WL}$, i.e. the capacity of WL test is the upper bound of the capacity of any aggregation-based GNN. Intuitively, with the metric to evaluate the network power, we further define the training process as the problem of optimizing the network capacity.

\textbf{Definition 4.} Let a GNN $\mathcal{A} = \mathcal{R} \circ \mathcal{L}^{(L)} \circ ... \circ \mathcal{L}^{(1)}$ with a fixed injective readout function $\mathcal{R}$, $G_1$ and $G_2$ are i.i.d. The training process for $\mathcal{A}$ is formulated as:
\begin{equation} \label{eq6}
\begin{split}
& \mathcal{L}^{(L)*}, ..., \mathcal{L}^{(1)*} = \\
& \max_{\mathcal{L}^{(L)}, ..., \mathcal{L}^{(1)}} \medspace \mathrm{Prob}\{ \mathcal{R} \circ \mathcal{L}^{(L)} \circ ... \circ \mathcal{L}^{(1)}(G_1) \\
& \neq \mathcal{R} \circ \mathcal{L}^{(L)} \circ ... \circ \mathcal{L}^{(1)}(G_2) | G_1 \ncong G_2 \}.
\end{split}
\end{equation}
Therefore, when training the network, the optimizer tries to find the best layer mapping for GNN to map non-isomorphic graphs into different embeddings as much as possible. With training process in Definition 4, we formulate the greedy layer-wise training for $\mathcal{A}$ as:
\begin{equation} \label{eq7}
\begin{split}
& \mathcal{L}^{(1)*} = \max_{\mathcal{L}^{(1)}} \medspace \mathrm{Prob}\{ \mathcal{R} \circ \mathcal{L}^{(1)}(G_1) \neq \mathcal{R} \circ \mathcal{L}^{(1)}(G_2) | G_1 \ncong G_2 \}, \\
& \mathcal{L}^{(2)*} = \max_{\mathcal{L}^{(2)}} \medspace \mathrm{Prob}\{ \mathcal{R} \circ \mathcal{L}^{(2)} \circ \mathcal{L}^{(1)*}(G_1) \\
& \neq \mathcal{R} \circ \mathcal{L}^{(2)} \circ \mathcal{L}^{(1)*}(G_2) | G_1 \ncong G_2 \}, \\
& ... \\
& \mathcal{L}^{(L)*} = \max_{\mathcal{L}^{(L)}} \medspace \mathrm{Prob}\{ \mathcal{R} \circ \mathcal{L}^{(L)} \circ \mathcal{L}^{(L-1)*} \circ ... \circ \mathcal{L}^{(1)*}(G_1) \\
& \neq \mathcal{R} \circ \mathcal{L}^{(L)} \circ \mathcal{L}^{(L-1)*} \circ ... \circ \mathcal{L}^{(1)*}(G_2) | G_1 \ncong G_2 \},
\end{split}
\end{equation}

In the following theorem, we provide a sufficient condition for a network trained layer-wise \eqref{eq7} to achieve the same capacity, as one trained from end to end \eqref{eq6}.

\begin{figure*}[t]
\begin{center}
  \includegraphics[width=0.95\linewidth]{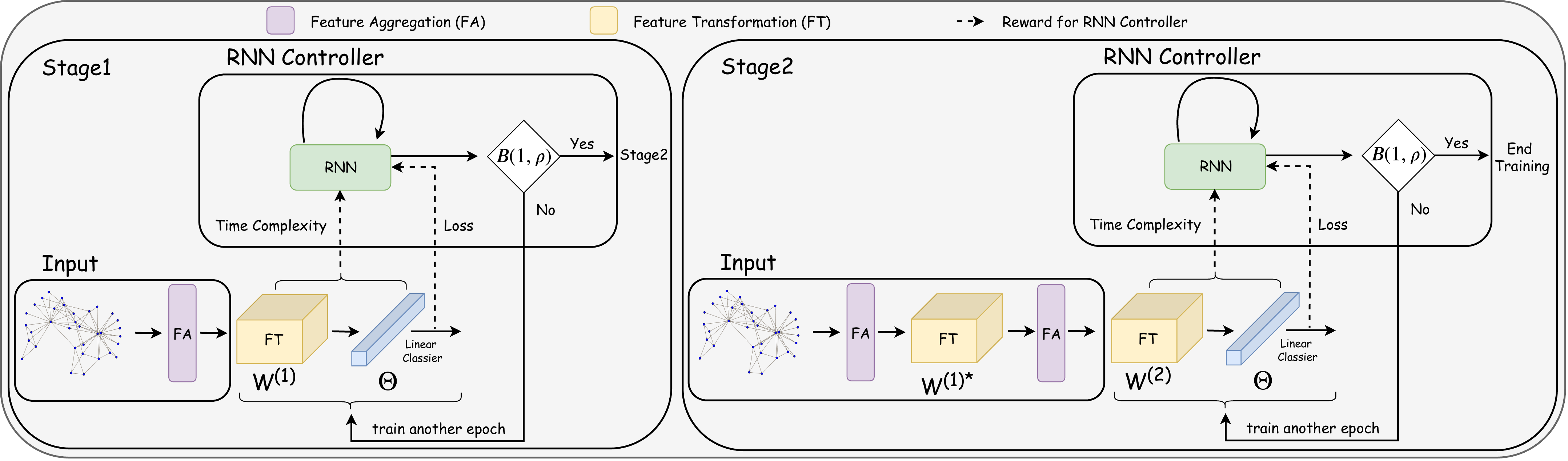}
\end{center}
  \caption{Learning to optimize in layerwise training algorithm for a two-layer GCN. Two layers are trained in two stage: in each stage the layer-wise trained network generates the training loss, the later index as the input for the RNN controller, and the RNN controller outputs the hidden stage for the next epoch, and the stopping probability $\rho$ for the current epoch, and sees whether the stopping probability $\rho$ is greater than the threshold probability $\rho_\mathrm{Thres}$. If $\rho > \rho_\mathrm{Thres}$ then the layer-wise training finishes, entering the next stage or ending the whole train (if it is the last layer), otherwise $\rho \le \rho_\mathrm{Thres}$ then training for another epoch. When one iteration of training process finishes, the reward of performance and efficiency will be fed back to the RNN controller, updating the weight and starting a new round of training.}
\label{figure2}
\end{figure*}

\textbf{Theorem 5.} Let $\mathcal{A} = \mathcal{R} \circ \mathcal{L}^{(L)} \circ ... \circ \mathcal{L}^{(1)}$ be a GNN with a fixed injective readout function $\mathcal{R}$. If $\mathcal{A}$ can be conventionally trained by solving the optimization problem \eqref{eq6} and the resulting $\mathcal{A}_{Con} = \mathcal{R} \circ \mathcal{L}^{(L)}_{Con} \circ ... \circ \mathcal{L}^{(1)}_{Con}$ is as powerful as the WL test given the conditions in Theorem 2, then $\mathcal{A}$ can also be layer-wise trained by solving the optimization problem \eqref{eq7} with the resulting $\mathcal{A}_{Lay} = \mathcal{R} \circ \mathcal{L}^{(L)}_{Lay} \circ ... \circ \mathcal{L}^{(1)}_{Lay}$ achieving the same capacity.


We provide the proof in the appendix. For the network architecture which is originally powerful enough through conventional training, we can train it to achieve the same capacity through layer-wise training. The idea of the proof is that: if there exists the injective mapping for each layer as the conditions in Theorem 2 satisfied, we can prove to find the injective mapping with layer-wise optimization problem as \eqref{eq7}. Otherwise, when the network architecture can not be powerful enough through conventional training, the following theorem establishes that the layer-wise trained network has non-decreasing capacity as the layer number increases. 



\textbf{Theorem 6.} Let GNN $\mathcal{A} = \mathcal{R} \circ \mathcal{L}^{(L)} \circ ... \circ \mathcal{L}^{(1)}$ with a fixed injective readout function $\mathcal{R}$, $G_1$ and $G_2$ are i.i.d., and $\boldsymbol{x}^{(l)}_i = \mathcal{L}^{(l)}(\boldsymbol{x}^{(l-1)}_i, \{ \boldsymbol{x}^{(l-1)}_j: j \in \mathcal{N}_i \}): \mathbb{R}^D \times \mathbb{M}^D \rightarrow \mathbb{R}^D$. With layer-wise training, if $\mathcal{L}^{(l)}_{Lay}$ is not guaranteed to be injective for $(\boldsymbol{x}^{(l-1)}_i, \{ \boldsymbol{x}^{(l-1)}_j: j \in \mathcal{N}_i \})$, but it still can distinguish different $\boldsymbol{x}^{(l-1)}_i$, i.e. if $\boldsymbol{x}^{(l-1)}_i \neq \boldsymbol{x}^{(l-1)}_k$, then $\mathcal{L}_{Lay}^{(l)}(\boldsymbol{x}^{(l-1)}_i, \{ \boldsymbol{x}^{(l-1)}_j: j \in \mathcal{N}_i \}) \neq \mathcal{L}_{Lay}^{(l)}(\boldsymbol{x}^{(l-1)}_k, \{ \boldsymbol{x}^{(l-1)}_j: j \in \mathcal{N}_k \})$, then we have that the capacity of the network is monotonically non-decreasing with deeper layers:
\begin{equation} \label{theorem6}
\begin{split}
& \mathrm{Prob}\{ \mathcal{R} \circ \mathcal{L}^{(l-1)}_{Lay} \circ ... \circ \mathcal{L}^{(1)}_{Lay}(G_1) \\
& \neq \mathcal{R} \circ \mathcal{L}^{(l-1)}_{Lay} \circ ... \circ \mathcal{L}^{(1)}_{Lay}(G_2) | G_1 \ncong G_2 \} \\
& \le \mathrm{Prob}\{ \mathcal{R} \circ \mathcal{L}^{(l)}_{Lay} \circ \mathcal{L}^{(l-1)}_{Lay} \circ ... \circ \mathcal{L}^{(1)}_{Lay}(G_1) \\
& \neq \mathcal{R} \circ \mathcal{L}^{(l)}_{Lay} \circ \mathcal{L}^{(l-1)}_{Lay} \circ ... \circ \mathcal{L}^{(1)}_{Lay}(G_2) | G_1 \ncong G_2 \} .
\end{split}
\end{equation}
We again direct readers to the appendix for the proof. The theorem indicates that, if the network architecture is not powerful enough through conventional training
, we can try to increase its capacity through training a deeper network. Layer-wise training can also train deeper GCNs more efficiently compared to state-of-the-arts.

What remains challenging is that the network capacity $C_{\mathcal{A}}$ is not available in an  analytical form with regards to network parameters.  In this study, we use the cross entropy as the loss function in classification tasks. More development in loss functions would be needed in future.

\subsection{L$^2$-GCN: Training with Learn Controllers}
One challenge to apply the layerwise training algorithm to graph convolutional networks (L-GCN) is that one may need to manually adjust the training epochs for each layer. A possible solution is early stopping, nevertheless it does not intuitively work well in L-GCN since the training loss in each layer is not comparable with the final validation loss. Motivated by learning to optimize \cite{andrychowicz2016learning, chen2017learning, li2016learning,cao2019learning}, we propose L$^2$-GCN, training a learned RNN controller to decide when to stop in each layer's training via policy-based REINFORCE \cite{williams1992simple}. The algorithm is illustrated in Figure \ref{figure2}. 

Specifically, we model the training process for L-GCN as a Markov Decision Process (MDP) defined as follows: i) \textbf{Action.} \medspace The action $a_t$ at time $t$ for the RNN controller is making the decision on whether to stop at the current-layer training or not. ii) \textbf{State.} \medspace The state $s_t$ at time $t$ is the loss in the current epoch, the layer index, and the hidden state of the RNN controller at time $t-1$. iii) \textbf{Reward.} \medspace The purpose of the RNN controller is to train the network efficiently with competitive performance, and therefore the non-zero reward is only received at the end of the MDP as the weighted sum of final loss and total training epochs (Time Complexity). iv) \textbf{Terminal.} \medspace Once the L-GCN finishes the $L$-layer training, the process terminates.

\begin{figure}[ht]
\begin{center}
  \includegraphics[width=1\linewidth]{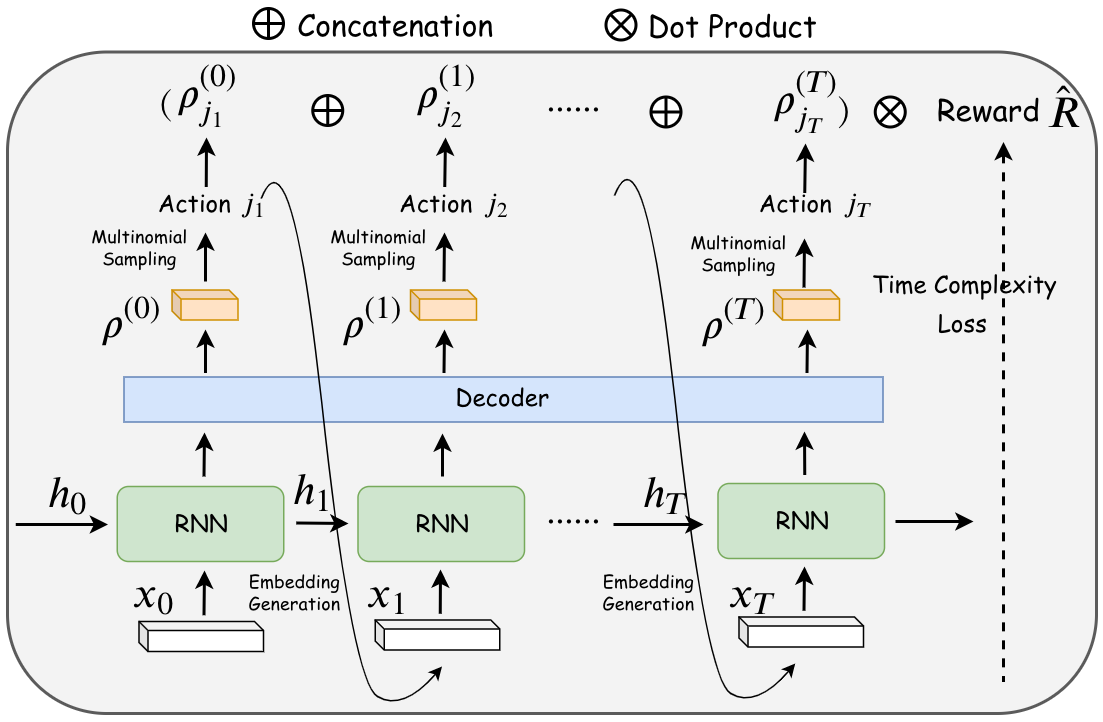}
\end{center}
  \caption{For each time step t, the RNN controller will first take the previous hidden vector $h_t$ and the generated embedding $x_t$ as input
  . Embedding $x_t$ is a concatenation of action embeddings, loss in the current epoch and the layer index. (Following \cite{gong2019autogan}, we generate embeddings randomly from a multinomial distribution for each action.) Then the RNN will output a hidden vector $h_{t+1}$, which will be decoded and classified by its corresponding softmax classifier. After obtaining probability vector $\rho^{(t)}$, it performs multinomial sampling to pick the action $j_{t+1}$ according to $\rho^{(t)}$. The relative probability $\rho^{(t)}_{t+1}$ will be further concatenated and recorded for reward calculation. When terminal, a final reward which is the dot product between picked action’s probability vector and $\hat{R}$ ($\hat R$ is the weighted sum of the final loss and total training epochs), will be generated for the controller to update its weights.}  
\label{fig:rnncontroller}
\end{figure}

Given the above settings, a sample trajectory from MDP will be: $( s_1, a_1, r_1, ..., s_t, a_t, r_t )$. The detailed architechture of RNN controller is shown in Figure \ref{fig:rnncontroller}. For each time step, the RNN will output a hidden vector, which will be decoded and classiﬁed by its corresponding softmax classiﬁer. The RNN controller works in an autogressive way, where the output of the last step will be fed into the next step. L-GCN will be sampled for each time step's output to decide whether to stop or not. When terminated, a final reward will be fed to the controller to update the weight.

\section{Experiments}
In this section, we evaluate predictive performance,  training time, and GPU memory usages of our proposed L-GCN and L$^2$-GCN on single- and multi-class classification tasks for six increasingly larger datasets: Cora \& PubMed \cite{kipf2016semi}, PPI \& Reddit \cite{hamilton2017inductive}, and Amazon-670K \& Amazon-3M \cite{chiang2019cluster}, as summarized in appendix. For Amazon-670K \& Amazon-3M, we use principal component analysis \cite{hotelling1933analysis} to reduce the feature dimension down to 100, and use the top-level categories as the class labels. The train/validate/test split is following the conventional setting for the inductive supervised learning scenario. We implemented our proposed algorithm in PyTorch \cite{paszke2017automatic}: for layer-wise training, we use the Adam optimizer with learning rate of 0.001 for Cora \& PubMed, and 0.0001 for PPI, Reddit, Amazon-670K and Amazon-3M; for RNN controller, we set the controller to make a stopping-or-not decision each 10 epochs (5 for Cora and  50 for PPI), use the controller architecture as in \cite{gong2019autogan} and the Adam optimizer with the learning rate of 0.05. All the experiments are conducted on a machine with GeForce GTX 1080 Ti GPU (11 GB memory), 8-core Intel i7-9800X CPU (3.80 GHz) and 16 GB of RAM. 


\subsection{Comparison with State of the Arts}
To demonstrate the efficiency and performance of our proposed algorithms, we compare them\ with state-of-the-arts in Table \ref{tab:sota}.  We compare L-GCN and L$^2$-GCN to the state-of-the-art GCN mini-batch training algorithms as GraphSAGE \cite{hamilton2017inductive}, FastGCN \cite{chen2018fastgcn} and VRGCN \cite{chen2017stochastic}, using their originally released codes and published settings, except that the batchsize and the embedding dimension of hidden layers are kept the same in all methods to ensure fair comparisons.  Specifically, we set the batchsize at 256 for Cora and 1024 for others; and we did the embedding dimension of hidden layers at 16 for Cora \& PubMed, 512 for PPI and 128 for others.
We do not compare the controller with other hyper-parameter tuning methods since the controller is widely used in many fields such as neural architecture search \cite{gong2019autogan}.

\begin{table*}[!htb]
\scriptsize
\begin{center}
\caption{Comparison with state-of-the-art on performance, training time and GPU memory usage (GPU memory usage during training). The best results for each row / dataset are highlighted in \textcolor{red}{red}.
}
\label{tab:sota}
\resizebox{1\textwidth}{!}{
\begin{tabular}{c | c c c | c c c | c c c | c c c | c c c}
\hline
\multirow{2}{*}{} & \multicolumn{3}{c|}{GraphSAGE \cite{hamilton2017inductive}} & \multicolumn{3}{c|}{FastGCN \cite{chen2018fastgcn}} & \multicolumn{3}{c}{VRGCN \cite{chen2017stochastic}} \multirow{2}{*}{} & \multicolumn{3}{|c|}{L-GCN} & \multicolumn{3}{c}{L$^2$-GCN} \\ \cline{2-16}
 & F1 (\%) & Time & Memory & F1 (\%) & Time & Memory & F1 (\%) & Time & Memory & F1 (\%) & Time & Memory & F1 (\%) & Time & Memory \\
\hline
Cora & 85.0 & 18s & 655M & \red{85.5} & 6.02s & 659M & 85.4 & 5.47s & \red{253M} & 84.7 & 0.45s & 619M & 84.1 & \red{0.38s} & 619M \\
PubMed & 86.5 & 483s & 675M & \red{87.4} & 32s & 851M & 86.4 & 118s & \red{375M} & 86.8 & 2.93s & 619M & 85.8 & \red{1.50s} & 631M \\
PPI & 68.8 & 402s & 849M & - & - & - & \red{98.6} & 63s & 759M & 97.2 & 49s & \red{629M} & 96.8 & \red{26s} & 631M \\
Reddit & 93.4 & 998s & 4343M & 92.6 & 761s & 4429M & \red{96.0} & 201s & 1271M & 94.2 & 44s & \red{621M} & 94.0 & \red{34s} & 635M \\
Amazon-670K & 83.1 & 2153s & 849M & 76.1 & 548s & 1621M & \red{92.7} & 534s & 625M & 91.6 & 54s & \red{601M} & 91.2 & \red{30s} & 613M \\
Amazon-3M & - & - & - & - & - & - & 88.3 & 2165s & 625M & \red{88.4} & 203s & \red{601M} & 88.4 & \red{125s} & 613M \\
\hline
\end{tabular}}
\end{center}
\end{table*}

On four common datasets Cora, PubMed, PPI and Reddit, we demonstrate that our proposed algorithm L-GCN is significantly faster than state-of-the-arts, with a consistent usage of GPU memory not dependent on dataset size, 
while maintaining  comparable  prediction  performance. With a learned controller to make the stopping decision, L$^2$-GCN can further reduce the training time (here we do not include search time) by half with tiny performance loss compared to L-GCN. For super large datasets, GraphSAGE and FastGCN fail to converge on Amazon-670K, and exceed the time limit on Amazon-3M in our experiment, whereas VRGCN achieves good performances after long training. Our proposed algorithms still stably achieve comparable performances efficiently on both Amazon-670K and Amazon-3M.

We did not include in Table \ref{tab:complexity} the time spent on hyper-parameter tuning (search) for any algorithm. Such a comparison was impossible as search time was not accessible for pre-trained state-of-the-arts.  Although a typical controller learning can be expensive (as reported in Table~\ref{tab:transferability}), RNN controllers in L$^2$-GCN learned over (especially large) datasets can be “transferrable” (shown next); and L$^2$-GCN without controller retraining actually saves time compared to dataset-specific manual tuning.  As to the memory usage,  the trends in practical GPU memory usages during training did not entirely agree with those in the theoretical analyses (Table~\ref{tab:complexity}). We contemplate that it is more likely in implementation: other models were implemented on TensorFlow and ours on PyTorch; and possible CPU memory usage of some models was unclear.



\subsection{Ablation Study}

\textbf{Transferability.}
We explore the transferability of the learned controller. Results in Table~\ref{tab:transferability} show that the controller learned from larger datasets could be reused for smaller ones (with similar loss functions) 
and thus save search time.

\begin{table}[ht]
\scriptsize
\begin{center}
\caption{The transferability of the learned contorllers.}
\label{tab:transferability}
\resizebox{0.45\textwidth}{!}{
\begin{tabular}{c | c c c | c c c}
\hline
 & \multicolumn{3}{c |}{Cora} & \multicolumn{3}{c}{PubMed} \\ \cline{2-7}
 & F1 (\%) & Train & Search & F1 (\%) & Train & Search \\
\hline
Controller-Cora & 84.1 & 0.38s & 16s & - & - & - \\
Controller-PubMed & 84.3 & 0.36s & 0s & 85.8 & 1.50s & 125s \\
Controller-Amazon-3M & 84.8 & 0.43s & 0s & 86.3 & 2.43s & 0s \\
\hline
\end{tabular}}
\end{center}
\end{table}

\begin{figure}[ht]
\begin{center}
  \includegraphics[width=1.0\linewidth]{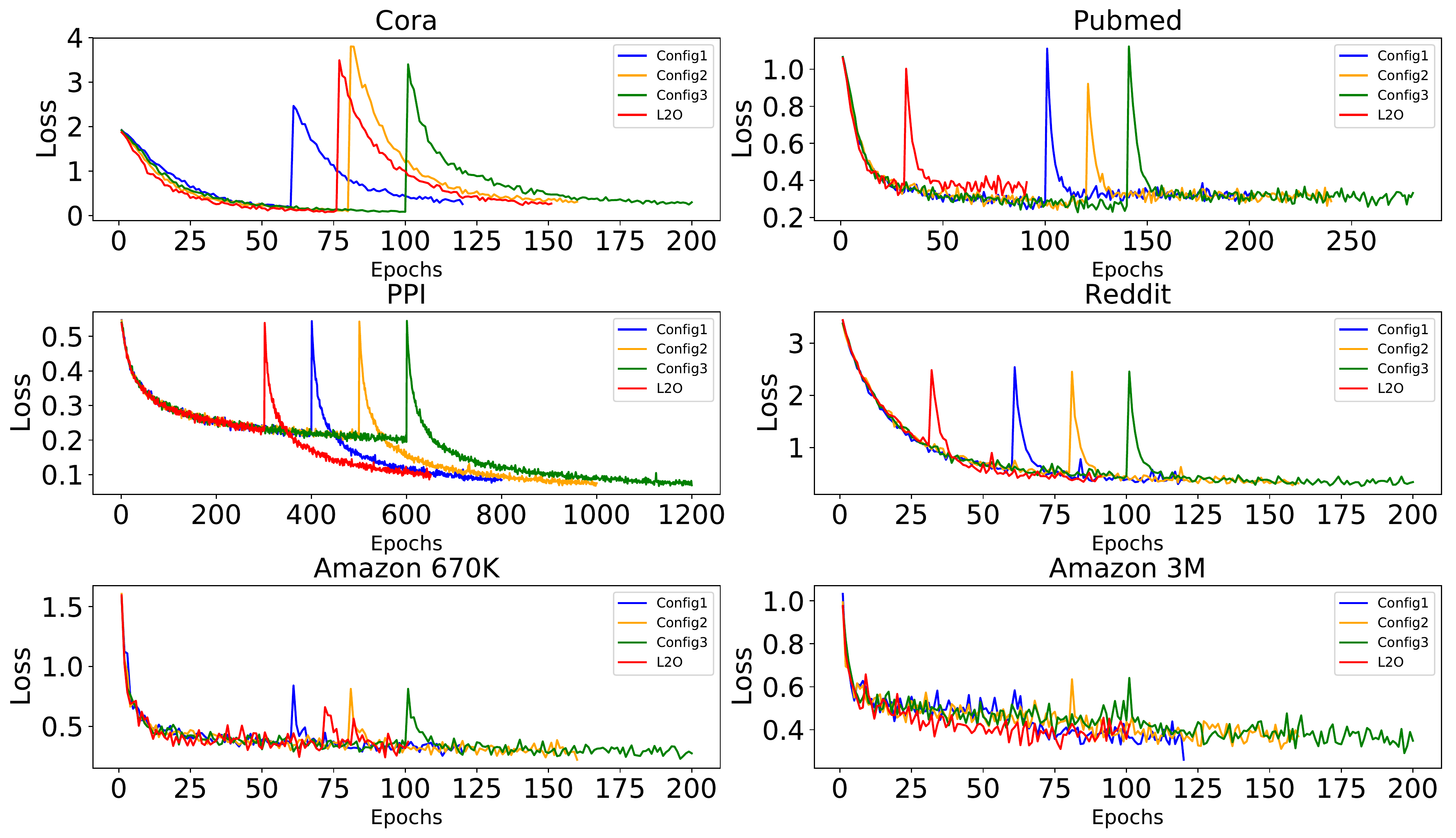}
\end{center}
  \caption{The loss curves for manual epoch configurations and the learning-to-optimize configuration.}
\label{fig:loss_curve}
\end{figure}

\textbf{Epoch configuration.} We consider the influence of different epoch configurations in layer-wise training on performance on six datasets.  Table \ref{tab:ablation-LW} shows that training under different epoch numbers in different layers  will affect the final performance. For layer-wise training (L-GCN), we configure different numbers of epochs for the two layers of our GCN as reported in Table \ref{tab:ablation-LW}. For layer-wise training with learning to optimize (L$^2$-GCN), we let the RNN controller to learn the epoch configuration from randomly sampled subgraphs as training data and report the automatically learned epoch numbers. Experimental results show that, trained with more epochs for each layer, L-GCN improves perfoemance except for Cora.
Moreover, with learning to optimize, the RNN controller in L$^2$-GCN automatically learns  epoch configurations with tiny performance loss but much less epochs. Figure \ref{fig:loss_curve} compares the loss curves of layer-wise training under various configurations and over various datasets.

\begin{table}[ht]
\scriptsize
\begin{center}
\caption{The influence of epoch configuration.}
\label{tab:ablation-LW}
\resizebox{0.46\textwidth}{!}{
\begin{tabular}{c | c c | c c | c c}
\hline
 & \multicolumn{2}{c |}{Cora} & \multicolumn{2}{c |}{PubMed} & \multicolumn{2}{c}{PPI} \\ \cline{2-7}
 & F1 (\%) & Epoch & F1 (\%) & Epoch & F1 (\%) & Epoch \\
\hline
L-GCN-Config1 & 83.2 & 60+60 & 86.8 & 100+100 & 93.7 & 400+400 \\
L-GCN-Config2 & 84.7 & 80+80 & 86.3 & 120+120 & 94.1 & 500+500 \\
L-GCN-Config3 & 83.0 & 100+100 & 86.4 & 140+140 & 94.9 & 600+600 \\
L$^2$-GCN & 84.1 & 75+75 & 85.8 & 30+60 & 94.1 & 300+350 \\
\hline
\hline
 & \multicolumn{2}{c |}{Reddit} & \multicolumn{2}{c |}{Amazon-670K} & \multicolumn{2}{c}{Amazon-3M} \\ \cline{2-7}
 & F1 (\%) & Epoch & F1 (\%) & Epoch & F1 (\%) & Epoch \\
\hline
L-GCN-Config1 & 93.0 & 60+60 & 91.4 & 60+60 & 88.2 & 60+60 \\
L-GCN-Config2 & 93.5 & 80+80 & 91.6 & 80+80 & 88.4 & 80+80 \\
L-GCN-Config3 & 93.8 & 100+100 & 91.7 & 100+100 & 88.3 & 100+100 \\
L$^2$-GCN & 92.2 & 30+60 & 91.2 & 70+30 & 88.0 & 20+80 \\
\hline
\end{tabular}}
\end{center}
\end{table}

\textbf{Deeper networks.} We evaluate the necessity of training a deeper network using layer-wise training. Previous attempts  seem to suggest the usefulness of training deeper GCN \cite{kipf2016semi}. However, the datasets used in the experiments there are not large enough to draw a definite conclusion.  Here we conduct experiments on 3 large datasets PPI, Reddit and Amazon-3M, with monotonically increasing layer number and total training epochs (each layer is trained for the same number of epochs) as shown in Table \ref{tab:ablation-layerNum}. Experimental results  show that, with more network layers, the prediction performance of layer-wise training gets better. Compared with 2-layer network, 4-layer L-GCN gains performance increase of 4.0, 0.7, and 0.8 (\%) on PPI, Reddit and Amazon-3M, respectively. When it comes to learn to optimize, the RNN controller learns a more efficient epoch configuration, while still achieving comparable performances as manually set epoch configurations.

\begin{table}[ht]
\scriptsize
\begin{center}
\caption{The influence of layer number on final performance.}
\label{tab:ablation-layerNum}
\resizebox{0.48\textwidth}{!}{
\begin{tabular}{c c | c c | c c | c c}
\hline
 & & \multicolumn{2}{c}{PPI} & \multicolumn{2}{|c}{Reddit} & \multicolumn{2}{|c}{Amazon-3M} \\ \cline{3-8}
 & & F1 (\%) & Epoch & F1 (\%) & Epoch & F1 (\%) & Epoch \\
\hline
 \multirow{2}{*}{2-layer} & L-GCN & 93.7 & 800 & 93.8 & 200 & 88.4 & 160 \\
 & L$^2$-GCN & 94.1 & 750 & 92.2 & 90 & 88.0 & 100 \\
\hline
 \multirow{2}{*}{3-layer} & L-GCN & 97.2 & 1200 & 94.2 & 300 & 89.0 & 240 \\
 & L$^2$-GCN & 96.8 & 650 & 94.0 & 210 & 88.7 & 120 \\
\hline
 \multirow{2}{*}{4-layer} & L-GCN & 97.7 & 1600 & 94.5 & 400 & 89.2 & 320 \\
 & L$^2$-GCN & 97.3 & 1100 & 94.3 & 250 & 89.0 & 170 \\
\hline
\end{tabular}}
\end{center}
\end{table}

Therefore, in layer-wise training, we have shown  that deeper layer networks can have better empirical performances, consistent with the theoretical, non-decreasing network capacity of deeper networks shown in Theorem 6.

\textbf{Applying layer-wise training to N-GCN.}
We also apply layer-wise training to N-GCN~\cite{abu2018n}, a  recent GCN extension. It consists of several GCNs over multiple scales so layer-wise training is applied to each GCN individually. Results in Table \ref{tab:ngcn} show that with layer-wise training, N-GCN is significantly faster with comparable performance.
\begin{table}[ht]
\scriptsize
\begin{center}
\caption{N-GCN with layer-wise training on Cora.}
\label{tab:ngcn}
\begin{tabular}{c | c c | c c}
\hline
 & \multicolumn{2}{c |}{Conventional Training} & \multicolumn{2}{c}{Layer-Wise Training} \\ \cline{2-5}
 & F1 (\%) & Time & F1 (\%) & Time \\
\hline
N-GCN & 83.6 & 62s & 83.1 & 4s \\
\hline
\end{tabular}
\end{center}
\end{table}





\section{Conclusions}
In this paper, we propose novel and efficient layerwise training algorithms for GCN (L-GCN) which separate feature aggregation and feature transformation during training and greatly reduce the complexity. Besides, we analyze theoretical grounds to rationalize the power of L-GCN in the graph isomorphism framework, provide a sufficient condition that L-GCN can be as powerful as conventional training, and prove that L-GCN is increasingly powerful as networks get deeper with more layers. Numerical results further support our theoretical analysis:  
our proposed algorithm L-GCN is significantly faster than state-of-the-arts, with a consistent usage of GPU memory not dependent on dataset size, while maintaining comparable prediction performance. Finally, motivated by learning to optimize, we propose L$^2$-GCN, designing an RNN controller to make the stopping decision for each-layer training and training it to learn to make the decision rather than manually configure the training epochs.  With the learned controller to make the stopping decision, L$^2$-GCN on average further reduces the training time by half with tiny performance loss,
compared to L-GCN.






{\small
\bibliographystyle{ieee_fullname}
\bibliography{ref}

\begin{thebibliography}{10}\itemsep=-1pt

\bibitem{abu2018n}
Sami Abu-El-Haija, Amol Kapoor, Bryan Perozzi, and Joonseok Lee.
\newblock N-{GCN}: Multi-scale graph convolution for semi-supervised node
  classification.
\newblock {\em arXiv preprint arXiv:1802.08888}, 2018.

\bibitem{andrychowicz2016learning}
Marcin Andrychowicz, Misha Denil, Sergio Gomez, Matthew~W Hoffman, David Pfau,
  Tom Schaul, Brendan Shillingford, and Nando De~Freitas.
\newblock Learning to learn by gradient descent by gradient descent.
\newblock In {\em Advances in neural information processing systems}, pages
  3981--3989, 2016.

\bibitem{cao2019learning}
Yue Cao, Tianlong Chen, Zhangyang Wang, and Yang Shen.
\newblock Learning to optimize in swarms.
\newblock In {\em Advances in Neural Information Processing Systems}, pages
  15018--15028, 2019.

\bibitem{chen2018fastgcn}
Jie Chen, Tengfei Ma, and Cao Xiao.
\newblock Fast{GCN}: fast learning with graph convolutional networks via
  importance sampling.
\newblock {\em arXiv preprint arXiv:1801.10247}, 2018.

\bibitem{chen2017stochastic}
Jianfei Chen, Jun Zhu, and Le Song.
\newblock Stochastic training of graph convolutional networks with variance
  reduction.
\newblock {\em arXiv preprint arXiv:1710.10568}, 2017.

\bibitem{chen2017learning}
Yutian Chen, Matthew~W Hoffman, Sergio~G{\'o}mez Colmenarejo, Misha Denil,
  Timothy~P Lillicrap, Matt Botvinick, and Nando de Freitas.
\newblock Learning to learn without gradient descent by gradient descent.
\newblock In {\em Proceedings of the 34th International Conference on Machine
  Learning-Volume 70}, pages 748--756. JMLR. org, 2017.

\bibitem{chiang2019cluster}
Wei-Lin Chiang, Xuanqing Liu, Si Si, Yang Li, Samy Bengio, and Cho-Jui Hsieh.
\newblock Cluster-gcn: An efficient algorithm for training deep and large graph
  convolutional networks.
\newblock {\em arXiv preprint arXiv:1905.07953}, 2019.

\bibitem{gong2019autogan}
Xinyu Gong, Shiyu Chang, Yifan Jiang, and Zhangyang Wang.
\newblock Auto{GAN}: Neural architecture search for generative adversarial
  networks.
\newblock In {\em Proceedings of the IEEE International Conference on Computer
  Vision}, pages 3224--3234, 2019.

\bibitem{grover2016node2vec}
Aditya Grover and Jure Leskovec.
\newblock node2vec: Scalable feature learning for networks.
\newblock In {\em Proceedings of the 22nd ACM SIGKDD international conference
  on Knowledge discovery and data mining}, pages 855--864. ACM, 2016.

\bibitem{hamilton2017inductive}
Will Hamilton, Zhitao Ying, and Jure Leskovec.
\newblock Inductive representation learning on large graphs.
\newblock In {\em Advances in Neural Information Processing Systems}, pages
  1024--1034, 2017.

\bibitem{hotelling1933analysis}
Harold Hotelling.
\newblock Analysis of a complex of statistical variables into principal
  components.
\newblock {\em Journal of educational psychology}, 24(6):417, 1933.

\bibitem{jiang2019accelerating}
Angela~H Jiang, Daniel L-K Wong, Giulio Zhou, David~G Andersen, Jeffrey Dean,
  Gregory~R Ganger, Gauri Joshi, Michael Kaminksy, Michael Kozuch, Zachary~C
  Lipton, et~al.
\newblock Accelerating deep learning by focusing on the biggest losers.
\newblock {\em arXiv preprint arXiv:1910.00762}, 2019.

\bibitem{kipf2016semi}
Thomas~N Kipf and Max Welling.
\newblock Semi-supervised classification with graph convolutional networks.
\newblock {\em arXiv preprint arXiv:1609.02907}, 2016.

\bibitem{lecun1995convolutional}
Yann LeCun, Yoshua Bengio, et~al.
\newblock Convolutional networks for images, speech, and time series.
\newblock {\em The handbook of brain theory and neural networks},
  3361(10):1995, 1995.

\bibitem{li2016learning}
Ke Li and Jitendra Malik.
\newblock Learning to optimize.
\newblock {\em arXiv preprint arXiv:1606.01885}, 2016.

\bibitem{paszke2017automatic}
Adam Paszke, Sam Gross, Soumith Chintala, Gregory Chanan, Edward Yang, Zachary
  DeVito, Zeming Lin, Alban Desmaison, Luca Antiga, and Adam Lerer.
\newblock Automatic differentiation in pytorch.
\newblock 2017.

\bibitem{perozzi2014deepwalk}
Bryan Perozzi, Rami Al-Rfou, and Steven Skiena.
\newblock Deepwalk: Online learning of social representations.
\newblock In {\em Proceedings of the 20th ACM SIGKDD international conference
  on Knowledge discovery and data mining}, pages 701--710. ACM, 2014.

\bibitem{tang2015line}
Jian Tang, Meng Qu, Mingzhe Wang, Ming Zhang, Jun Yan, and Qiaozhu Mei.
\newblock Line: Large-scale information network embedding.
\newblock In {\em Proceedings of the 24th international conference on world
  wide web}, pages 1067--1077. International World Wide Web Conferences
  Steering Committee, 2015.

\bibitem{wang2019e2}
Yue Wang, Ziyu Jiang, Xiaohan Chen, Pengfei Xu, Yang Zhao, Yingyan Lin, and
  Zhangyang Wang.
\newblock E2-train: Training state-of-the-art cnns with over 80\% energy
  savings.
\newblock In {\em Advances in Neural Information Processing Systems}, pages
  5139--5151, 2019.

\bibitem{weisfeiler1968reduction}
Boris Weisfeiler and Andrei~A Lehman.
\newblock A reduction of a graph to a canonical form and an algebra arising
  during this reduction.
\newblock {\em Nauchno-Technicheskaya Informatsia}, 2(9):12--16, 1968.

\bibitem{williams1992simple}
Ronald~J Williams.
\newblock Simple statistical gradient-following algorithms for connectionist
  reinforcement learning.
\newblock {\em Machine learning}, 8(3-4):229--256, 1992.

\bibitem{wu2019simplifying}
Felix Wu, Tianyi Zhang, Amauri Holanda~de Souza~Jr, Christopher Fifty, Tao Yu,
  and Kilian~Q Weinberger.
\newblock Simplifying graph convolutional networks.
\newblock {\em arXiv preprint arXiv:1902.07153}, 2019.

\bibitem{xu2018powerful}
Keyulu Xu, Weihua Hu, Jure Leskovec, and Stefanie Jegelka.
\newblock How powerful are graph neural networks?
\newblock {\em arXiv preprint arXiv:1810.00826}, 2018.

\bibitem{ying2018graph}
Rex Ying, Ruining He, Kaifeng Chen, Pong Eksombatchai, William~L Hamilton, and
  Jure Leskovec.
\newblock Graph convolutional neural networks for web-scale recommender
  systems.
\newblock In {\em Proceedings of the 24th ACM SIGKDD International Conference
  on Knowledge Discovery \& Data Mining}, pages 974--983. ACM, 2018.

\bibitem{you2019drawing}
Haoran You, Chaojian Li, Pengfei Xu, Yonggan Fu, Yue Wang, Xiaohan Chen,
  Yingyan Lin, Zhangyang Wang, and Richard~G Baraniuk.
\newblock Drawing early-bird tickets: Towards more efficient training of deep
  networks.
\newblock {\em arXiv preprint arXiv:1909.11957}, 2019.

\bibitem{zhang2018link}
Muhan Zhang and Yixin Chen.
\newblock Link prediction based on graph neural networks.
\newblock In {\em Advances in Neural Information Processing Systems}, pages
  5165--5175, 2018.

\bibitem{zou2019layer}
Difan Zou, Ziniu Hu, Yewen Wang, Song Jiang, Yizhou Sun, and Quanquan Gu.
\newblock Layer-dependent importance sampling for training deep and large graph
  convolutional networks.
\newblock In {\em Advances in Neural Information Processing Systems}, pages
  11249--11259, 2019.

\end{thebibliography}
}

\end{document}